\pdfoutput=1

\documentclass[11pt]{article}

\usepackage[]{naacl2021}

\usepackage{times}
\usepackage{latexsym}
\usepackage{enumerate}

\usepackage[T1]{fontenc}

\usepackage[utf8]{inputenc}

\usepackage{microtype}

\title{Fine-grained Interpretation and Causation Analysis in Deep NLP Models}

\author{
Hassan Sajjad ~~~ Narine Kokhlikyan\textsuperscript{*} ~~~ Fahim Dalvi ~~~ Nadir Durrani \\~\\
{\tt \{hsajjad,faimaduddin,ndurrani\}@hbku.edu.qa} \\ 
Qatar Computing Research Institute, HBKU Research Complex, Doha 5825, Qatar \\
{\tt narine@fb.com} \\
\textsuperscript{*}Facebook AI, 1 Facebook Way, Menlo Park, CA 94025, USA \\ 

}

\begin{document}
\maketitle
\section{Introduction}
Deep neural networks have constantly pushed the state-of-the-art performance in natural language processing (NLP) and are now considered as the de facto modeling approach in solving most complex NLP tasks such as machine translation, summarization and question-answering. Despite the benefits and the usefulness of deep neural networks at-large, their opaqueness is a major cause of concern. 
Interpreting neural networks is considered important for increasing trust in AI systems, providing additional information to decision makers, and assisting ethical decision making~\cite{lipton2016mythos}.

Work done on neural interpretation can be broadly divided into two types: i) intrinsic analysis, ii) causation analysis. The former class 
analyze representations learned within neural networks 
~\cite{DBLP:journals/corr/AdiKBLG16,belinkov:2017:acl,belinkov-etal-2017-evaluating,dalvi:2017:ijcnlp,conneau-etal-2018-cram,tenney-etal-2019-bert,durrani-etal-2019-one}.
The latter 
tries to understand the network with various behavioural studies 
~\cite{gulordava-etal-2018-colorless,linzen_tacl,marvin-linzen-2018-targeted}, 
or by studying importance of input features and neurons with respect to a  prediction~\cite{DBLP:journals/corr/abs-1805-12233,shappely_NIPS2017_7062,tran2018importance}. These works on interpretation have been presented in earlier tutorials, for example ACL 2020 \cite{belinkov-etal-2020-interpretability} and EMNLP 2020 \cite{wallace-etal-2020-interpreting}. The former focused on representation analysis and behavioral studies and the latter discusses work done on the causal interpertation such as assessing model's behavior using constructed examples. Both these great tutorials serve as a starting point for the new researchers in this area.


The representation analysis, also called as structural analysis, is useful to understand how various core linguistic properties are learned in the model \cite{belinkov-etal-2020-linguistic}. However, the analysis suffers from a few limitations. It  mainly focuses at interpreting full vector representations and does not study the role of fine-grained components in the representation i.e. neurons. Also the findings of representation analysis do not link with the cause of a prediction~\cite{belinkov-glass-2019-analysis}. While the behavioral analysis evaluates model predictions, it does not typically connect them with the influence of the input features and the internal components of a model~\cite{vig_causal_genderbias_2020}. 

In this tutorial, we 
present and discuss the research work on interpreting 
fine-grained components of a model 
%
%
from two perspectives, i) fine-grained interpretation, ii) causation analysis. The former 
introduces methods to analyze individual neurons and a group of neurons with respect to a 
language property or a task. The latter 
studies the role of neurons and input features in explaining decisions made by the model.
We will cover important research questions such as i) how is knowledge distributed across the model components? ii) what knowledge learned within the model is used for specific predictions? iii) does the inhibition of specific knowledge in the model change predictions? iv) how do different modeling and optimization choices impact the underlying knowledge?

Recent work on interpreting neurons has shown that in-addition to gaining better understanding of the inner workings of neural networks, the neuron-level interpretation has applications in model distillation \cite{rethmeier2019txray}, domain adaptation \cite{gu2021pruningthenexpanding} or efficient feature selection \cite{dalvi-2020-CCFS} e.g., by removing unimportant neurons, facilitating architecture search, and mitigating model bias by identifying neurons responsible for sensitive attributes like gender, race or politeness \cite{bau2018identifying,vig_causal_genderbias_2020}. 
These recent works are not only enabling better understanding of these networks, but are also leading towards better, fairer and more environmental-friendly models, which are all important goals for the 
Artificial Intelligence community at large. 

\section{Description}
\label{sec:description}

The tutorial 
is divided into two main parts: i) fine-grained interpretation, and ii) causation analysis. 
%
%
The first part of the tutorial
covers  methods that align neurons to human interpretable concepts or study the most salient neurons in the network. We cluster these methods into four groups i) Visualization Methods \cite{karpathy2015visualizing,li-etal-2016-visualizing}, ii) Corpus Selection \cite{kadar-etal-2017-representation,poerner-etal-2018-interpretable,Na-ICLR, Mu-Nips}, iii) Neuron Probing \cite{dalvi:2019:AAAI,lakretz-etal-2019-emergence,valipur-2019,durrani-etal-2020-analyzing,durrani-FT} and iv) Unsupervised Methods \cite{bau2018identifying,torroba-hennigen-etal-2020-intrinsic,wu:2020:acl,michael-etal-2020-asking}. 
We 
also discuss evaluation methods 
that are used to 
measure the effectiveness of an interpretation method, such as accuracy, control tasks~\cite{hewitt-liang-2019-designing} and ablation studies~\cite{LiMJ16a,DBLP:journals/corr/abs-1812-05687,dalvi:2019:AAAI,lakretz-etal-2019-emergence}. 
Moreover, we 
cover various applications of these methods that go beyond interpretation such as efficient transfer learning~\cite{dalvi-2020-CCFS}, controlling system's behavior \cite{bau2018identifying, suau2020finding}, generating explanations ~\cite{Mu-Nips} and domain adaptation \cite{gu2021pruningthenexpanding}.

The second part, \textit{Causation Analysis}, 
focuses on the methods that 
characterize the role of neurons and layers towards a specific prediction. More concretely, we 
discuss gradient and perturbation-based attribution algorithms such as Integrated Gradients~\cite{ig}, Layer Conductance~\cite{cond}, Saliency~\cite{saliency}, SHapley Additive exPlanations(SHAP)~\cite{shappely_NIPS2017_7062} and showcase how they can help us to identify important neurons in different layers of a deep neural network.
Besides that we 
also dive deep into more recent and advanced attribution algorithms that take feature or neuron interactions into account. More specifically, we 
look into Integrated Hessians~\cite{janizek2020explaining}, Shapely Taylor index~\cite{dhamdhere2020shapley} and Archipelago~\cite{achipelago}.


Lastly, we 
present various open source toolkits and libraries that provide implementation of notable techniques in the area. A few examples  
include: Captum~\cite{kokhlikyan2020captum}, InterpretML\footnote{https://interpret.ml/}, NeuroX~\cite{neurox-aaai19:demo}, Ecco\footnote{https://www.eccox.io/} and Diagnnose~\cite{diagnnose}. We 
give a walk-through of how some of these tools can be used for fine-grained interpretation and causation analysis. 
Throughout the tutorial, 
we critically evaluate 
the strengths and weakness of 
the presented methods, 
and 
discuss future directions. 

%

\section{Outline}

\begin{enumerate}[I.]
    \item \textbf{Introduction:} We 
    introduce the topic and motivate it by providing the vision of model interpretability, and how it leads towards fair and ethical models 
    and towards better generalization. We 
    then describe various forms of interpretation and 
    outline the scope of the tutorial (15 minutes).
    
    
    \item{\textbf{Fine-grained Interpretation:}} We 
    present and discuss the work on neuron-level interpretation. 
    (90 minutes)
    
    \begin{itemize}
    \item{Methods} (30 minutes)
    \item{Evaluation} (15 minutes)
    \item{Findings} (30 minutes)
    \item{Practical} (15 minutes)
    \end{itemize}
    
    \item{\textbf{Causation Analysis:}} In causation analysis we 
    present various methods on interpreting model predictions with respect to input features and individual neurons. (60 minutes)
    
    \begin{itemize}
    \item{Methods} (30 minutes)
    \item{Evaluation} (10 minutes)
    \item{Practical} (20 minutes)
    \end{itemize}
    
    
    
    
    \item \textbf{Concept-based Interpretation of Prediction:} This part 
    aims to 
    bridge the gap between fine-grained interpretation and causation analysis. We 
    discuss how fine-grained interpretation and causation analysis can be combined to establish concept-based interpretation of model predictions. 
    (10 minutes)
    
    \item \textbf{Discussion:} The last part 
    discusses the overall challenges that the current work faces and suggests future directions. (10 minutes)

\end{enumerate}

\section{Prerequisites}
We assume a basic knowledge of the deep learning and familiarity with the LSTM-based and transformer-based pre-trained models such as ELMo~\cite{peters-etal-2018-deep} and BERT~\cite{devlin-etal-2019-bert}. Additionally, some familiarity with natural language processing tasks such as, named entity tagging, natural language inference, etc. would be 
useful but not mandatory. We do not expect participants to have familiarity with the research on the interpretation and analysis of deep models. 
Familiarity with Python, Pytorch and Transformers library~\cite{Wolf2019HuggingFacesTS} would be 
useful to understand the practical part.

\section{Reading List}
\begin{itemize}
    \item In order to get an overview of the interpretation field, trainees may look at the following survey papers: \newcite{belinkov-glass-2019-analysis} and \newcite{danilevsky-etal-2020-survey}.
    \item Fine-grained analysis and its Applications: \newcite{,bau2018identifying,dalvi:2019:AAAI,Mu-Nips,suau2020finding} etc.
    \item Causation analysis: \newcite{shappely_NIPS2017_7062} provides an overview of various methods introduced in literature. For more details, see the following papers: ~\newcite{voita2020analyzing,ig,cond,lime,janizek2020explaining}
\end{itemize}
In addition to the above list, interested trainees may look at the papers mentioned in Section \ref{sec:description}.

\section{Instructor Information (Alphabetic order}

\noindent\textbf{Fahim Dalvi}, Software Engineer, Qatar Computing Research Institute, Qatar

\noindent Email: faimaduddin@hbku.edu.qa \\
\noindent Website: \url{https://fdalvi.github.io}
\\~\\
Fahim Dalvi is an experienced Software Engineer with a demonstrated history of working in the research industry and is currently employed at the Qatar Computing Research Institute. Fahim's research is centered around the intersection of Natural Language Processing and Deep Learning, and he has worked on wide variety of problems in these fields including Machine Translation, Language Modelling and Explainability in Deep Neural Networks. He also spends his time converting research into practical applications, with a focus on scalable web applications. Fahim also spends some time every year mentoring and teaching Deep Learning at Fall and Summer schools.
\\~\\
\noindent\textbf{Hassan Sajjad}, Senior Research Scientist, Qatar Computing Research Institute, Qatar

\noindent Email: hsajjad@hbku.edu.qa \\
\noindent Website: \url{https://hsajjad.github.io}
\\~\\
Hassan Sajjad is a Senior Research Scientist at the Qatar Computing Research Institute (QCRI), HBKU. His research interests include the interpretation of deep neural models, machine translation, domain adaptation, and natural language processing involving low-resource and morphologically-rich languages. His research work has been published in several prestigious venues such as CL, CSL, ICLR, ACL, NAACL and EMNLP. His work in collaboration with MIT and Harvard on the interpretation of deep models has also been featured in several tech blogs including MIT News. Hassan co-organized BlackboxNLP 2020, and the WMT 2019/2020 machine translation robustness task. He served as an area chair for the analysis and interpretability, NLP Application, and machine translation tracks at various *CL conferences. In addition, Hassan has been regularly teaching courses on deep learning internationally at various spring and summer schools. 
\\~\\
\noindent\textbf{Narine Kokhlikyan}, Research Scientist, Facebook AI

\noindent Email: narine@fb.com \\
\noindent Website: \url{https://www.linkedin.com/in/narine-k-88916721/}
\\~\\
Narine is a Research Scientist focusing on Model Interpretability as part of PyTorch team at Facebook. Her research interests include the understanding of Deep Neural Network internals and their predictions across different applications such as Natural Language Processing, Computer Vision and Recommender Systems.  In the recent years she gave talks and presented tutorials  on  Model Interpretability  at KDD 2020 and NeurIPS 2019.
Before joining Facebook Narine worked on Natural Language Processing, Time Series Analysis and numerical optimizations.
\\~\\
\noindent\textbf{Nadir Durrani}, Senior Research Scientist, Qatar Computing Research Institute, Qatar \\
\noindent Email: ndurrani@hbku.edu.qa \\
\noindent Website: \url{http://alt.qcri.org/~ndurrani/}
\\~\\
Nadir Durrani is a Research Scientist at the Arabic Language Technologies group at Qatar Computing Research Institute. His research interests include interpretation of neural networks, neural and statistical machine translation (with focus on reordering, domain adaptation, transliteration, dialectal translation, pivoting, closely related and morphologically rich languages), eye-tracking for MT evaluation, spoken language translation and speech synthesis. His recent work focuses on analyzing contextualized representations with the focus of linguistic interpretation, manipulation, feature selection and model distillation. His work on analyzing deep neural networks has been published at venues like Computational Linguistics, *ACL, AAAI and ICLR. Nadir has been involved in co-organizing workshops such as simultaneous machine translation and WMT 2019/2020 Machine translation robustness task. He regularly serves as program committee and has served as Area chair at ACL and AAAI this year. 

\bibliography{anthology,eacl2021}
\bibliographystyle{acl_natbib}




\end{document}